\pdfoutput=1

\documentclass[11pt]{article}

\usepackage[]{acl}

\usepackage{times}
\usepackage{latexsym}

\usepackage[T1]{fontenc}

\usepackage[utf8]{inputenc}

\usepackage{microtype}
\usepackage{amsmath}
\usepackage{graphicx}
\usepackage{amsfonts}
\usepackage{multirow} 
\usepackage{makecell}
\title{Evidence-Enhanced Triplet Generation Framework for Hallucination Alleviation in Generative Question Answering}

\author{Haowei Du,  Huishuai Zhang,  Dongyan Zhao}

\begin{document}
\maketitle
\begin{abstract}
To
address the hallucination in generative question answering (GQA) where the answer can not be derived from the document, we propose a novel evidence-enhanced triplet generation framework,
EATQA, encouraging the model to
predict all the combinations of ⟨Question, Evidence, Answer⟩ triplet
by flipping the source pair and the target label
to understand their logical relationships, i.e.,
predict Answer(A), Question(Q), and Evidence(E) given a QE, EA, and QA
pairs, respectively. Furthermore, we bridge the distribution gap to distill the knowledge from evidence in inference stage. Our framework ensures the model to learn the logical relation between query, evidence and answer, which simultaneously improves the evidence generation and query answering. In this paper, we apply EATQA to LLama and it outperforms other LLMs-based methods and hallucination mitigation approaches on two challenging GQA benchmarks. Further analysis shows that our method not only keeps prior knowledge within LLM, but also mitigates hallucination and generates faithful answers. 
\end{abstract}

\section{Introduction}

Large language models (LLM) represent a significant milestone in the development of general artificial intelligence \cite{brown2020language, touvron2023LLama, chowdhery2023palm}. While these
models have demonstrated unprecedented performance across various general tasks, they still face a series of challenges, including
issues such as hallucination \cite{tonmoy2024comprehensive}
and handling long contexts \cite{jin2024llm}. In document-based generative question answering (GQA) \cite{lewis2018generative}, model may generate answers which are inconsistent with the document or mismatch the query, well known as hallucinations \cite{gunjal2024detecting, liu2024survey}.  
Recent works utilize an external model to retrieve relevant information and agument the factuality of generation. However, the intrinsic discrepancy between retriever and LLM may introduce information which is only surface relevant but not helpful to answer the question \cite{salemi2024evaluating}.

\begin{figure}[h]
    \centering
\includegraphics[width=0.5\textwidth]{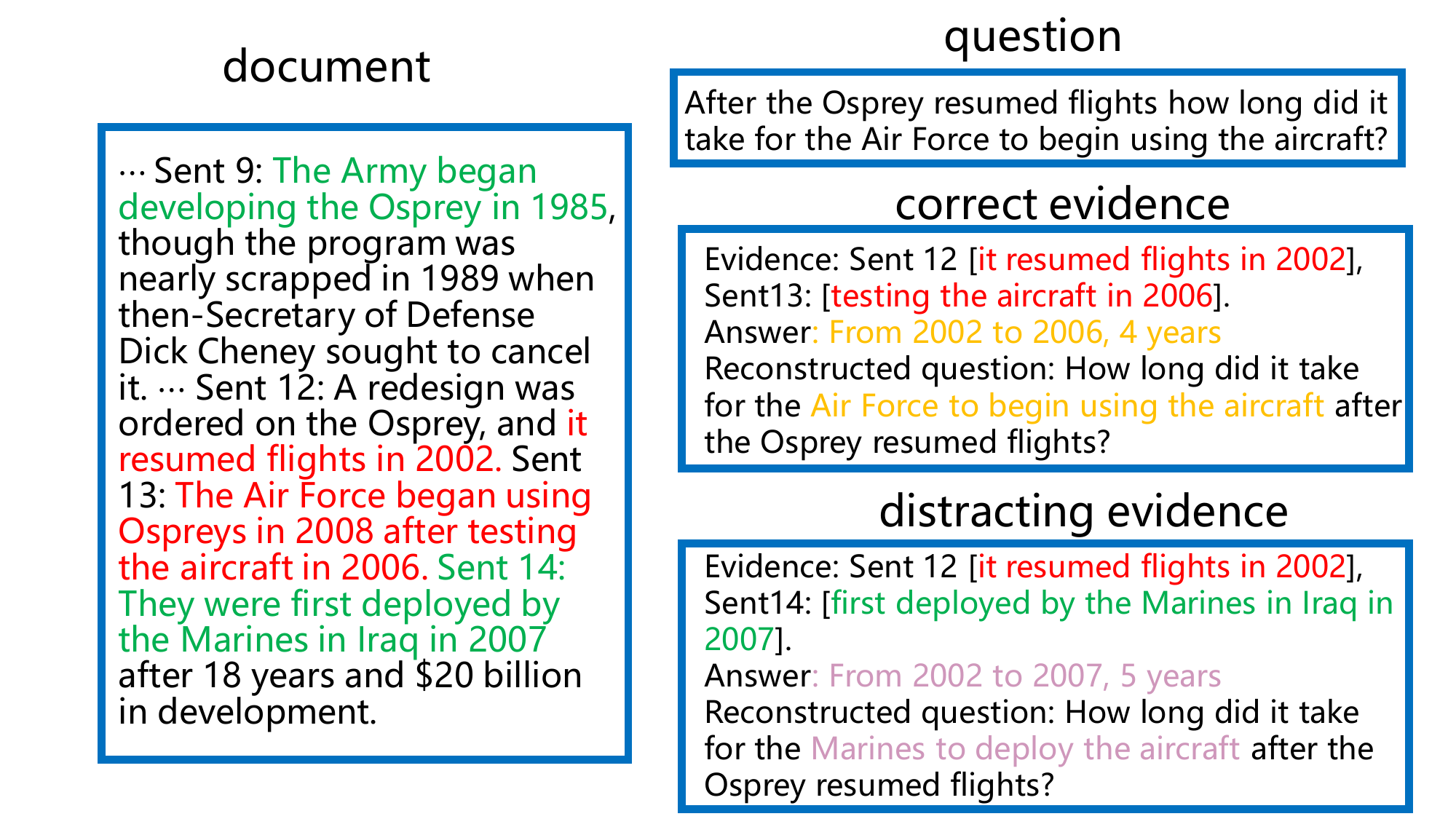}
    \caption{One example from MultiRC dataset. Red denotes supporting evidence and green denotes misleading sentences.}
    \label{case1}
\end{figure}

To enhance the logical reasoning and avoid the misleading information, we highlight the supporting evidence of the answer in document-based QA. Instead of the retrieve-then-read pipeline, we let the LLM to tackle the generation of evidence and answer in a unified triplet generation framework, where each two of <question, evidence, answer> serve as the input in the respective instruction to generate the other. The evidence is utilized to reconstruct the question to ensure the model captures its logical relations to question and answer instead of superficial relevance.

We take an example in MultiRC \cite{khashabi2018looking}  dataset in Figure \ref{case1}. The question is ``After the Osprey resumed flights how long did it take for the Air Force to begin using the aircraft?
'' and the answer can not be extracted from a single sentence in the document. To answer this question, the model needs to find the multiple evidence ``Osprey resumed flights in 2002''
and ``Air Force began using Ospreys in 2008 after testing the aircraft in 2006'', and decide the answer is ``4 years''. If the model is distracted by the incorrect evidence ``Marines develop the aircraft in Iraq in 2007'', it will derive the wrong answer ``5 years''. On the other hand, based on the correct evidence, the model can reconstruct the original question because the correct evidence contains the enough information. However, with the incorrect evidence, the reconstructed question is ``How long did it take for the Marines to begin using the aircraft $\cdots$ '' which is not consistent with the original question. It shows the correct evidence is crucial for the question answering and the reconstruction of the question based on evidence and answer can reflect the correctness of the evidence.

To alleviate the hallucination and enhance the logical reasoning between the question, evidence and answers, we propose our \MakeUppercase{\textbf{e}}vidence enh\MakeUppercase{\textbf{a}}nced \MakeUppercase{\textbf{t}}riplet generation framework (EATQA), which includes three instruction tuning tasks to predict all the combinations of ⟨Question, Evidence, Answer⟩ triplet
by flipping the source pair and the target label
to understand their logical relationships, i.e.,
predict A(Answer), Q(Question), and E(Evidence) given a QE, EA, and QA
pairs, respectively. We close the QA distribution distance between evidence-aware and evidence-absent settings (distribution bridging) to distill the knowledge from evidence to mitigate the gap in inference stage when the evidence sentences can not be derived.

We conduct experiments in two widespread document-based GQA datasets with diverse answer types, MultiRC and QASPER, based on different sizes of LLMs. Compared with different sizes of the backbone model, our unified triplet generation framework shows significant improvement on the two datasets, becoming the new state-of-the-art. Further analysis demonstrates the ability of our approach to tackle longer document with more sentences. Moreover, we observe the positive correlation among the performance of 3 subtasks in the triplet generation framework, which shows the effectiveness of unifying the generation of the three parts with one LLM in the triplet generation framework.

We conclude our contributions as follows:
\textbf{1.} We highlight the evidence retrieval to alleviate hallucinations of LLM in GQA task. Instead of utilizing another LM as the retriever, which may introduce misleading information, we propose the unified triplet generation framework including three instruction tuning tasks to improve the logical reasoning ability of LLM for GQA task. Specifically, we train the LLM to generate the Answer(A), Question(Q), and Evidence(E) based on the information of QE, EA, and QA
pairs respectively. The distribution bridging is utilized to distill knowledge from evidence for inference stage. 
\textbf{2.} We conduct experiments on two widespread document-based QA datasets MultiRC and QASPER with different sizes of LLM, and achieve the new state-of-the-art on both datasets. \textbf{3.} Further experiments show the effectiveness of unified triplet generation framework on both evidence retrieval and question answering. Moreover, our method not only keeps the prior knowledge within LLM, but also mitigates the hallucination for questions beyond the internal knowledge.

\section{Related Work}
Generative question answering (GQA) aims to generate an abstractive answer rather than extract an answer to a given question
from provided passages \cite{fan2019eli5, li2021addressing}.
Early works on GQA mostly tried to improve the faithfulness of the answer by investigating
reliable external knowledge sources or incorporating multiple information sources. \citet{yin2015neural}
propose Neural Generative Question Answering, an end-to-end model that generates
answers to simple factoid questions based on the knowledge base, while \citet{bi2019incorporating} propose the
Knowledge-Enriched Answer Generator (KEAG) to generate a natural answer by integrating facts
from four different information sources, namely, questions, passages, vocabulary, and knowledge.

Recent works focus more on the conditional generation model. 
\citet{li2021addressing}
propose Rationale-Enriched Answer Generator (REAG), in which they add an extraction task to
obtain the rationale for an answer at the encoding stage, and the decoder is expected to generate the answer based on both the extracted rationale and original input. \citet{su2022read} propose a framework named RBG (read before generate), to jointly models answer generation with machine reading. They augment the generation model with fine-grained, answer-related salient information predicted by the MRC module, to enhance answer faithfulness. Such methods can exploit and utilize the information in the original input better, while they require the extra effort of building models to extract that information. 
CAD \cite{shi2023trusting} follows a contrastive
output distribution that amplifies the difference between
the output probabilities when a model is
used with and without context.
RHO \cite{ji2023rho} introduce
local and global knowledge-grounding techniques
into dialogue generation and further utilize a conversational
reasoning model to re-rank the generated
responses.

Our approach differs from these methods in 3 folds:
1. Our model does not need the discourse structure including paragraphs and sections, and focus on the sentence-level evidence retrieval. 2. We tackle the logical relationship between question, evidence and answer in the unified triplet generation framework with the vanilla LLM, which prevents the intrinsic discrepancy of external retriever. 3. We claim correct evidence with answers contains the enough information to reconstruct the question, and utilize this characteristic to alleviate the generation of relevant but not needed evidence.

\section{Methodology}
\begin{figure}[h]
    \centering
\includegraphics[width=0.5\textwidth]{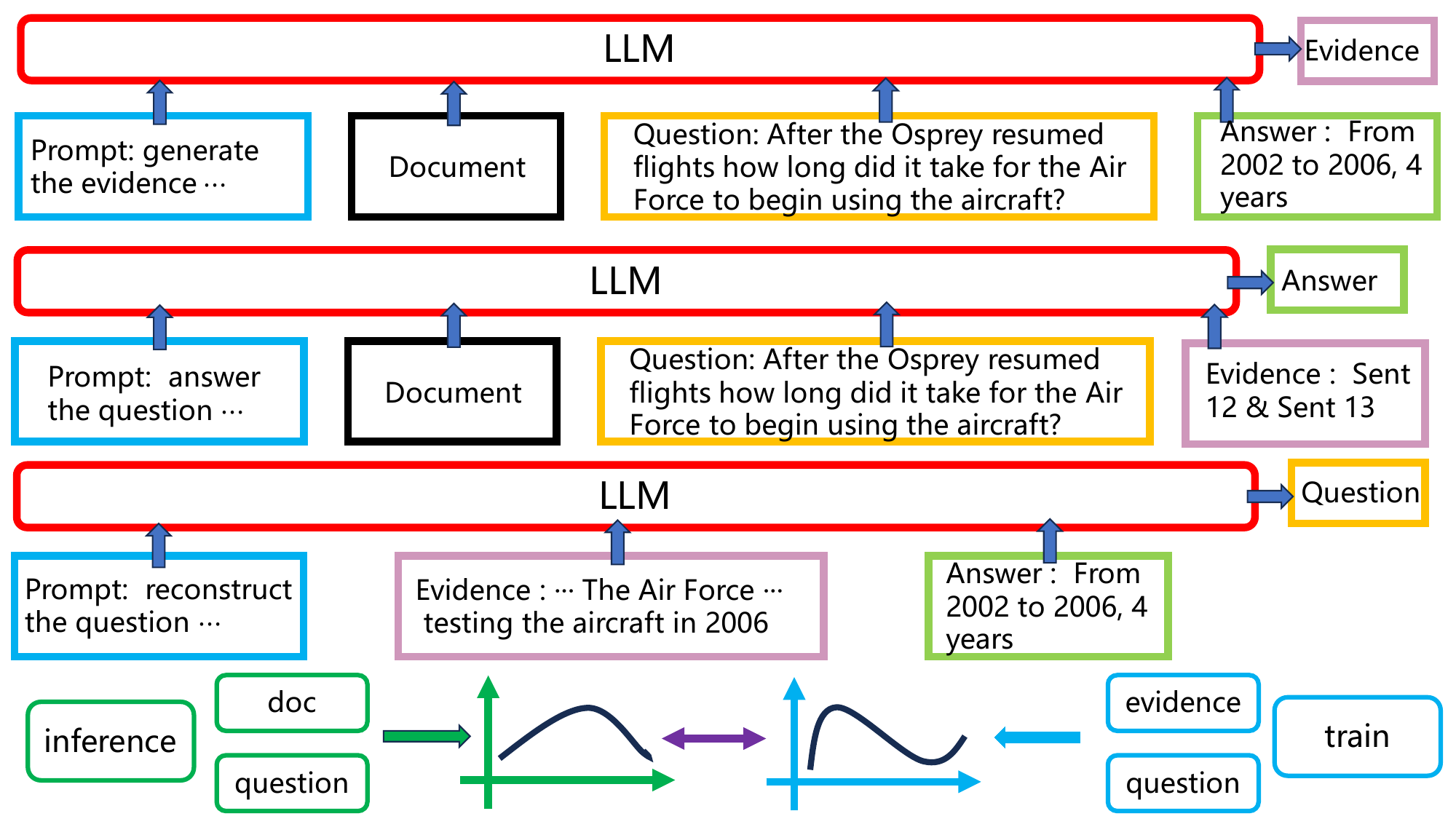}
    \caption{Model overview of EATQA.}
    \label{pipe}
\end{figure}

In this part, first we introduce the document-based GQA task and the model architecture of Triplet-QA. Then we present the unified triplet generation framework, which predicts all the combinations of ⟨Question, Evidence, Answer⟩ triplet
by flipping the source pair and the target label
to learn their logical relationships. 
Specifically, our method contains 3 subtasks, answer-aware evidence retrieval, evidence-aware query answering and evidence-aware query restoration, which is shown in Figure \ref{pipe} from upper to down.

The motivation of triplet generation framework comes from that the posterior  probability of question answering is positively proportional to the probability of evidence generation and question recovery by Bayesian formulation: 

\begin{equation}
\begin{aligned}
& \mathbf{P_M}(a |q,e,d)     = \frac{\mathbf{P_M}(a,q,e, d)} {\mathbf{P_M}{(q,e,d)}} \\
&=  \frac{\mathbf{P_M}(a,d) \mathbf{P_M}(e|a,d)  \mathbf{P_M}(q|e,a,d)} {\mathbf{P_M}{(q,e,d)}}
\end{aligned}
\end{equation}
where $d$, $q$, $e$, $a$ denote the document, question, evidence and answer. It shows the accuracy of QA should be proportional to the accuracy of evidence extraction and question restoration. We assume the evidence sentences contain the sufficient information to reconstruct the question, i.e. $\mathbf{P_M}(q|e,a)$ = $\mathbf{P_M}(q|e,a,d)$.

As the basis of our framework, we explain the feasibility of our method. Take query restoration as an example: in Figure \ref{case1}, only from the answer ``4 years'', the model is hard to restore the query because there may be multiple sentences involving ``4 years'' in the document. However, given the evidence sentences which point to the key events ``Osprey resumed flights'' and ``Air Force begin using the aircraft'', the model has derived the basic part of the query and it enables our query restoration module reasonable, which improves the model ability to arrange the information to restore the query.

\subsection{Preliminary}

Document-base GQA task aims to generate the answer of the natural text question based on the document including multiple sentences:
\begin{align}
    D = [s_1, s_2, \cdots, s_n]
\end{align}
where $D$ denotes the document, $s_i$ denotes the i-th sentence in the document and $n$ denotes the number of sentences.
The model can be formulated as a function of \begin{equation}
\mathbf{f_M(a)}=\prod_{i=1}^{n} P(a_i | a_1, a_2, \cdots, a_{i-1}, \mathbf{q}, \mathbf{D})
\end{equation}
where $n$ denotes the answer length, $q$ denotes the query and $a_0$ denotes the begin-of-speech (BOS) token.
Generally the answer has flexible forms which can not be directly extracted from the document.

\subsection{Model Architecture}
EATQA is built on the widespread LLM, LLama \cite{touvron2023LLama} with a few additional
learnable parameters. we additionally adopt several trainable adapter
tokens $p = [p_1, p_2, \cdots, p_{N_p} ]$ which are prepended
to the key and value of each self-attention layer,
where $N_p$ is the number of adapter tokens.
So the number of trainable parameters
of EATQA 7B is 4.5M, only 0.06\% of total
parameters of LLama 7B. With such a few
trainable parameters, EATQA effectively preserves
LLMs’ prior knowledge and the casual reasoning ability to 
understand the logical relations between the question, evidence and answer.
EATQA consists of three objectives: answer-aware evidence generation, evidence-enhanced query answering and evidence-aware query restoration. 

\subsubsection{Answer-Aware Evidence Generation (QA->E)}
In this part, we model the probability of supporting evidence extraction for the query-answer pair $\mathbf{P_M}(e|a,d)$. We design the instruction and ask the LLM to generate the evidence which supports the query and the corresponding answers. Therefore, the input to model is the instruction, source document, the query and the corresponding answer. The output of model is the supporting evidence. The specific instruction is ``generate the relevant evidence from the document to answer the following question'' and we insert the document, question and answers into the template in Figure \ref{temp}.

As for the loss function, by Bayesian Formula \cite{mises1942correct} we derive
\begin{equation}
\begin{aligned}
& \log(P(e,q))  = \log \int P(e,q,a) d_a \\
 &= \log \int q(a|e,q) \frac{P(e,q,a)}{q(a|e,q)}d_a \\
 &\geq \int q(a|e,q) \log( \frac{P(e,q,a)}{q(a|e,q)})d_a \\ 
 &=  E_{q(a|e,q)} \log( \frac{P(e,q,a)}{q(a|e,q)}) \\
 &= E_{q(a|e,q)} \log(\frac{P(a,q) P(e|a,q)}{q(a|e,q)}) \\
 &= E_{q(a|e,q)} \log(P(e|a,q)) + E_{q(a|e,q)} \log(\frac{P(a,q)}{q(a|e,q)}) \\
 &=  E_{q(a|e,q)} \log(P(e|a,q)) - KL(P(a,q) || q(a|e,q)) 
\end{aligned}
\label{kl}
\end{equation}
where $q(a|e,q)$ denotes the probability of answer $a$ to the question $q$ holds based on the evidence $e$, which is produced by the same backbone in our method with specific prompt, $KL$ denotes Kullback-Leibler divergence \cite{van2014renyi}. To maximize the evidence extraction probability, we should maximize the probability of evidence supporting the question-answer pair $P(e|a,q)$ and minimize the distribution distance between question answering with or without evidence $KL(P(a,q) || q(a|e,q))$. Considering we can not get access to the golden evidence during inference stage, the second term $KL(P(a,q) || q(a|e,q)) $, named as ``\textbf{distribution bridging}'', narrows down the gap between training and inference for answer generation. 
we utilize cross-entropy loss function to optimize the probability $P(e|a,q)$:
\begin{equation}
\begin{aligned}
\mathcal{L}_{\mathrm{QAE}} & =-\log P(\mathbf{e} \mid \mathbf{D}, \mathbf{q}, \mathbf{a}) \\
& =-\sum_{t=0}^{N_e-1} \log P\left(e_{t+1} \mid \mathbf{D}, \mathbf{q}, \mathbf{a}, e_{\leq t}\right)
\end{aligned}
\end{equation}
where $D$ denotes the document, $N_e$ denotes the length of the evidence, $P\left(e_1 \mid \mathbf{D}, \mathbf{q}, \mathbf{a}, e_{\leq 0}\right):=P\left(e_1 \mid \mathbf{D}, \mathbf{q}, \mathbf{a}\right)$.


\subsubsection{Evidence-Enhanced Question Answering (QE->A)}
In this part, we ask the LLM to generate the answers based on the corresponding question and the relevant evidence. The instruction is ``generate the correct answers for the following question based on the document and the evidence support the answers to the question.'', and we insert the instruction, document, question and evidence into the template in Figure \ref{temp}, as inputs into the LLM. 
The objective function formulated as:
\begin{equation}
\begin{aligned}
\mathcal{L}_{\mathrm{seq}} & =-\log P(\mathbf{a} \mid \mathbf{D}, \mathbf{q}, \mathbf{e}) \\
& =-\sum_{t=0}^{N_a-1} \log P\left(a_{t+1} \mid \mathbf{D}, \mathbf{q}, \mathbf{e}, a_{\leq t}\right)
\end{aligned}
\end{equation}
where $N_a$ denotes the length of the answers, $P\left(a_1 \mid \mathbf{D}, \mathbf{q}, \mathbf{e}, a_{\leq 0}\right):=P\left(a_1 \mid \mathbf{D}, \mathbf{q}, \mathbf{e}\right)$.
This task can be seen as the main task of EATQA and enables the model to derive the answers based on the question and evidence.
On the other hand, to narrow the gap between training and inference, we minimize second term of Eq.\ref{kl}: $KL(P(a,q) || q(a|e,q))$. When the evidence provided is incomplete or has misleading information, the model learns to resort to the original document for the answer, which improves the robustness of training stage. 
Therefore, the loss function of this part is:
\begin{equation}
\begin{aligned}
\mathcal{L}_{\mathrm{QEA}} & = \mathcal{L}_{\mathrm{Seq}} + \alpha_{kl} \cdot \mathbf{KL}(\mathbf{P}(a,q) || \mathbf{q}(a|e,q)) \\
\end{aligned}
\end{equation}
where $\alpha_{kl}$ denotes the hyper-parameter to tune.

\subsubsection{Evidence-Aware Question Restoration (EA->Q)}
In this part, we aim to model the probability of $\mathbf{P_M}(q|e,a)$ and ask the LLM to recover the question based on the evidence-answer pair. The prompt is ``reconstruct the question based on the answers and corresponding supporting evidence'', and we insert the prompt, document, evidence and answers into the template in Figure \ref{temp}.
The objective function is formulated as:
\begin{equation}
\begin{aligned}
\mathcal{L}_{\mathrm{EAQ}} & =-\log P(\mathbf{q} \mid \mathbf{D}, \mathbf{e}, \mathbf{a}) \\
& =-\sum_{t=0}^{N_q-1} \log P\left(q_{t+1} \mid \mathbf{D}, \mathbf{e}, \mathbf{a}, q_{\leq t}\right)
\end{aligned}
\end{equation}
where $N_q$ denotes the length of the question, $P\left(q_1 \mid \mathbf{D}, \mathbf{e}, \mathbf{a}, q_{\leq 0}\right):=P\left(q_1 \mid \mathbf{D}, \mathbf{e}, \mathbf{a}\right)$.
Considering the incorrect evidence does not contain the full information of the original question, this objective helps to enhance the casual relations between evidence and answers.

\begin{table}
\centering
\begin{tabular}{lccc}
\hline Methods & EM  & F1 & \#Para.\\
\hline
GPT-3 (32 shot) & 30.5 & 75.4 & 175B \\
Flan-T5 & - & 83.4 & 137B \\
T5 & 63.1 & 88.1 & 11B \\
Vega-2   & 62.4 & 88.2 &6B  \\
 ERNIE-3.0 & 63.2 & 88.6 & 10B \\
PALM  &63.6 & 88.7 & 540B \\ 
\hline
LLama2-7B & 57.2 &  86.1 &  7B \\
RAG & 58.1 &  86.7 &  7B\\
CAD & 58.2 &  87.2 &  7B\\
RHO & 59.4 &  87.3 &  7B\\
EAT-QA-7B & 61.9 & 88.1 & 7B \\ 
\hline
LLama2-13B & 62.0 &  87.9 &  13B \\
RAG & 63.1 &  88.1 &  13B\\
CAD & 63.5 &  88.3 &  13B\\
RHO & 64.2 &  88.4 &  13B\\
 EATQA-13B & \textbf{65.5} & \textbf{89.8} & 13B \\ 
\hline
\end{tabular}
\caption{Results on MultiRC dataset compared with competitive LLM methods. ``\#Para.'' denotes the parameter number in the model. GPT-3 reports few shot results with 32 examples in the prompt without parameter updating, and other methods are finetuning results from the original paper. We conduct 5 experiments with different random seeds and our method significantly beats the prior SOTA, with p-value less than 0.001.}
\label{result-multi}
\end{table}

\begin{table}
\centering
\begin{tabular}{lcc}
\hline Methods & F1 & \#Para.\\
\hline 
VCC-3B & 38.2 & 3B \\
SE-Mistral-7B & 39.3 & 7B \\
TOVA-7B & 42.0 & 7B \\
ChatGLM3-6B-32k & 43.3 & 6B \\
\hline
LLama2-7B-PI & 42.4 &  7B \\
RAG & 43.9 & 13B\\
CAD & 43.1 &  13B\\
RHO & 43.2 &  13B\\
EATQA-7B & \textbf{45.1} & 7B \\ 
\hline
\end{tabular}
\caption{Results on Qasper dataset compared with competitive LLM methods. ``\#Para.'' denotes the parameter number in the model. The results are cited from the original paper. LLama2-7B-PI denotes using Position Interpolation with LLama2-7B.}
\label{result-qasper}
\end{table}

\subsection{Training and Inference}
With the unified optimization of all three EATQA objectives, our model captures the logical relations between question, evidence and answers. The overall objective is the weighted accumulation: 
\begin{equation}
L_{Triplet} = \alpha_1 L_{QAE} + \alpha_2 L_{QEA} + \alpha_3 L_{EAQ}
\label{train}
\end{equation}
where $\alpha_1$, $\alpha_2$ and $\alpha_3$ are tuneable hyper-parameters.

Because of the design of distribution bridging, we do not need to first generate the evidence based on the question and then construct QE->A template in the inference stage. Instead, we directly instruct the model to generate the answer from the original document, which keeps the inference efficiency.


\section{Experiments}

\begin{table}
\centering
\begin{tabular}{lccc}
\hline Methods & EM  & F1 & \#Para.\\
\hline 
\multicolumn{4}{c}{w/ LLama2-7B} \\
\hline
backbone & 57.2 &  86.1 &  7B \\
-Question Restoration & 60.2 & 87.1 & 7B \\ 
-Evidence Generation  &60.8 & 87.7 & 7B \\ 
-KL  &61.0 & 87.6 & 7B \\ 
\hline
EATQA-7B & 61.9 & 88.1 & 7B \\
\hline
\multicolumn{4}{c}{w/ LLama2-13B} \\
\hline
backbone &  62.0 &  87.9 &  13B \\
-Query Restoration & 64.2 & 88.7 & 13B \\ 
-Evidence Generation  &64.5 & 88.6 & 13B \\ 
-KL & 64.6 & 89.1 & 13B \\ 
 EATQA-13B & 65.5 & 89.8 & 13B \\ 
\hline
\end{tabular}
\caption{Ablation results with LLama2 from 7B to 13B on MultiRC dataset. }
\label{ablation}
\end{table}

\subsection{Datasets}
We evaluate  on two widespread benchmark GQA datasets, MultiRC \cite{khashabi2018looking} and QASPER \cite{dasigi2021dataset}, across different
domains.
MultiRC creates multi-domain multi-hop questions, where documents across various domains are selected from multiple datasets.
Each instance consists of a document including about 15 sentences. All instances were constructed
such that it is not possible to answer a question
correctly without gathering information from multiple sentences.
Qasper includes 5049 questions over 1585 Natural Language Processing papers in the academic
research domain focusing on entire papers, which is designed to facilitate document-grounded, information-seeking QA. Qasper contains a variety of answer types,
including extractive, abstractive, yes/no, and unanswerable questions.

We utilize Exact Match (EM) and F1 scores \cite{opitz2019macro} to evaluate our method. The F1 score measures the overlap of answer tokens between the predicted and ground-truth answer. EM is more strict which awards point if any of the annotated answers is generated exactly. 

\subsection{Implementation Details}
We conduct experiments with LLama2 \cite{touvron2023LLama} from 7B to 13B as the LLM. To reduce computation cost and keep prior knowledge in LLM, we use LoRA \cite{hu2021lora}, which freezes the pretrained model weights and injects trainable rank decomposition matrices into each
layer of the LLM. The parameters $\alpha_1$, $\alpha_2$, and $\alpha_3$ in Eq. \ref{train} are tuned from [0.1, 0.3, 0.5, 0.7, 1.0], and set to 0.3, 1.0 and 0.3 in our method. We use AdamW as optimizer and the initial learning rate is set to 3e-5. 
Because the maximum input length of LLama2 is 4096 and the average context length of QASPER is about 16K, we utilize position interpolation \cite{chen2023extending} to extend the context length to 32K.

\subsection{Baselines}
We compare our method with existing widespread LLMs including T5-11B \cite{raffel2020exploring}, Flan-137B \cite{wei2021finetuned}, Vega2-6B \cite{zhong2022toward}, GPT-3 (few shot) \cite{brown2020language}, LoRAMoE \cite{dou2023loramoe}, PaLM 540B \cite{anil2023palm} for MultiRC. 
For Qasper, we compare our method with LLM-based long context methods, AttenWalker \cite{nie2023attenwalker}, ChatGLM3-6B-32k \cite{du2021glm}, SE-Mistral-7B \cite{jin2024llm}, VCC-3B \cite{zeng2024vcc} and TOVA-7B \cite{oren2024transformers}
For hallucination mitigation methods, we compare our approach against RAG with Dense Passage Retriever (DPR) \cite{karpukhin2020dense}, CAD \cite{shi2023trusting}, RHO \cite{ji2023rho} using the same backbone.

\subsection{Results}
From Table \ref{result-multi}, compared with the backbone, our method improves by 3.7 EM and 2.0 F1 on 7B-scale model as well as 2.3 EM and 1.5 F1 on 13B-scale model. It demonstrates the effectiveness of our evidence enhanced triplet generation framework on document based GQA. Moreover, our method with 13B parameters outperforms the 540B PaLM finetuning by 0.7 EM and 1.0 F1, becoming the new state-of-the-art. Our method with 7B-scale model has achieved the comparable performance on F1 with larger models like T5-xxl and ERNIE-3.0. 

From Table \ref{result-qasper}
 compared with the backbone, our method improves by 2.7 F1 on 7B-scale model. Qasper contains more rigorous samples and existing hallucination mitigation methods struggle to improve the performance. It demonstrates the effectiveness of our method on challenging long document QA.

\section{Analysis}

\begin{table}
\centering
\begin{tabular}{lcc}
\hline
Probability & LLama2 & EATQA \\
 \hline
$P(Y_{A|Q} = \hat{Y})$ & 34.8 & 37.1 \\
$P(Y_{A|Q,D} = \hat{Y} | Y_{A|Q} = \hat{Y})$ & 88.8 &  85.8 \\
$P(Y_{A|Q,D} = \hat{Y} | Y_{A|Q} \neq  \hat{Y})$ & 48.7 & 52.2 \\
\hline
\end{tabular}
\caption{Prior knowledge mitigation and hallucination mitigation. $Y_{A|Q}$ denotes the answer generated based on the vanilla query by QA model, which reflects the prior knowledge of LLM. $Y_{A|Q,D}$ denotes the answer generated based on the query and document. $\hat{Y}$ denotes the golden answer. }
\label{hallucination}
\end{table}

\begin{figure*}[h]
	
	\begin{minipage}{0.3\linewidth}
		\centerline{\includegraphics[width=\textwidth]{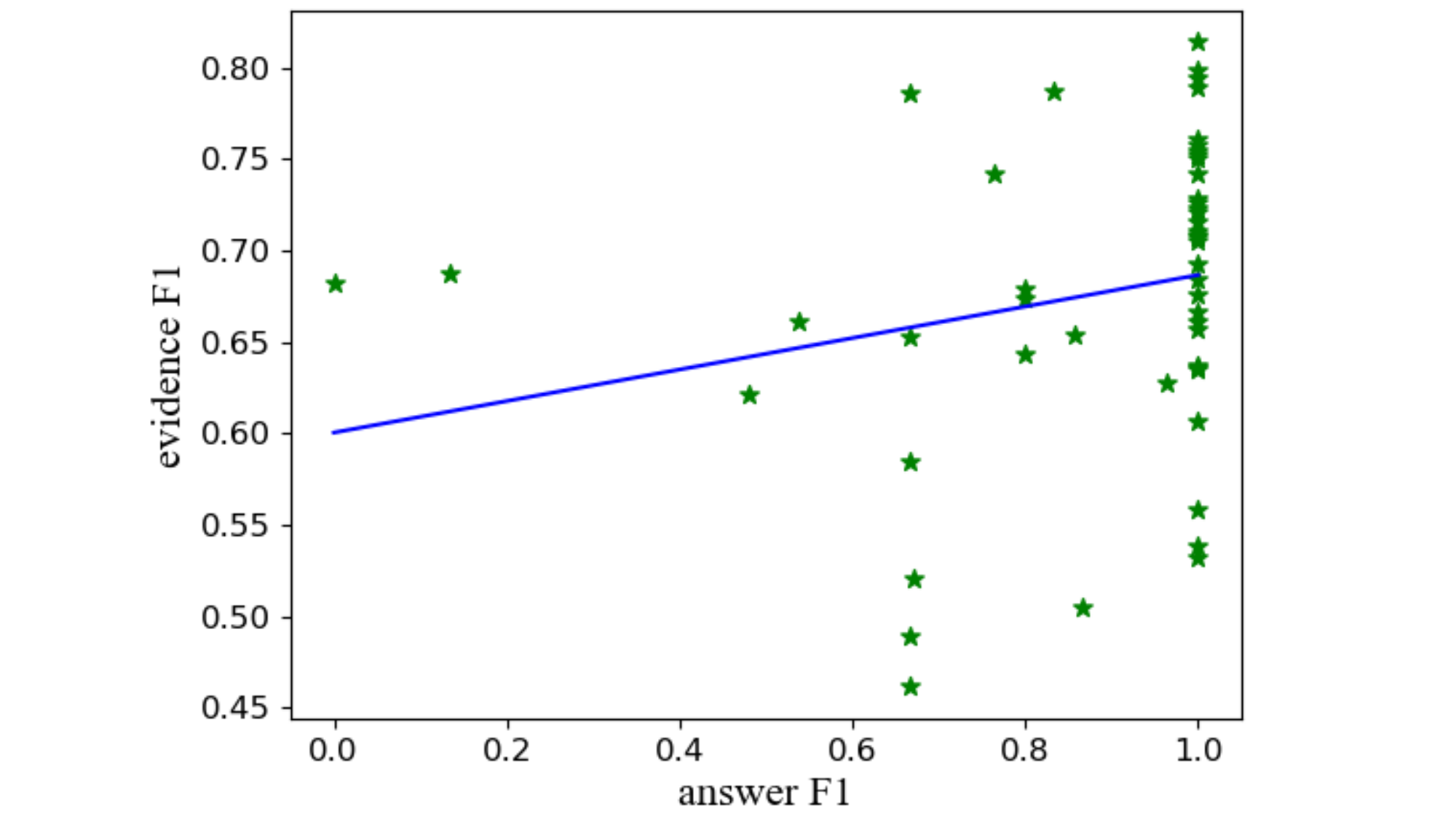}}
		\centerline{QEA vs QAE}
	\end{minipage}
	\begin{minipage}{0.3\linewidth}
		\centerline{\includegraphics[width=\textwidth]{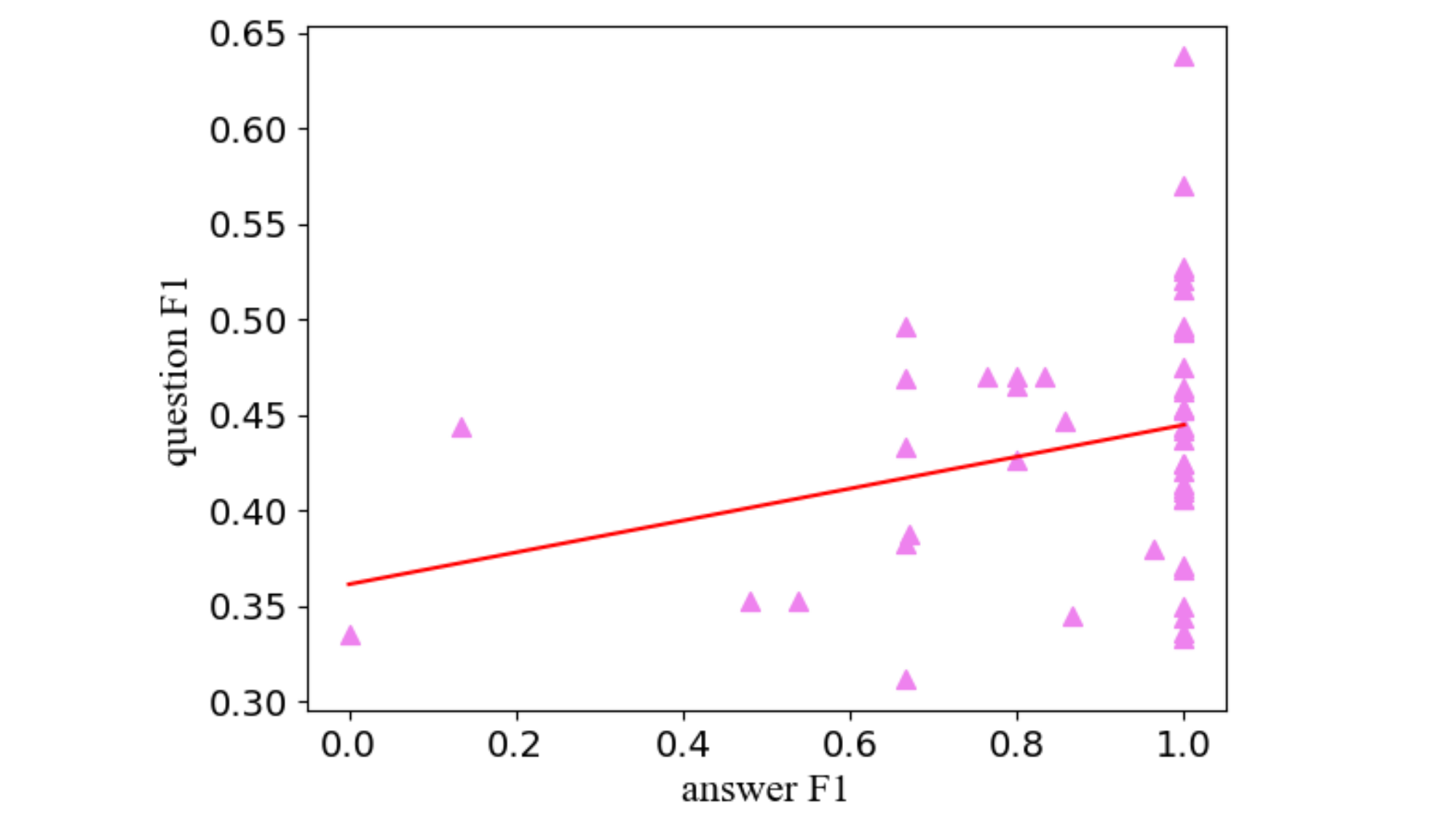}}
	 
		\centerline{QEA vs EAQ}
	\end{minipage}
	\begin{minipage}{0.3\linewidth}
		\centerline{\includegraphics[width=\textwidth]{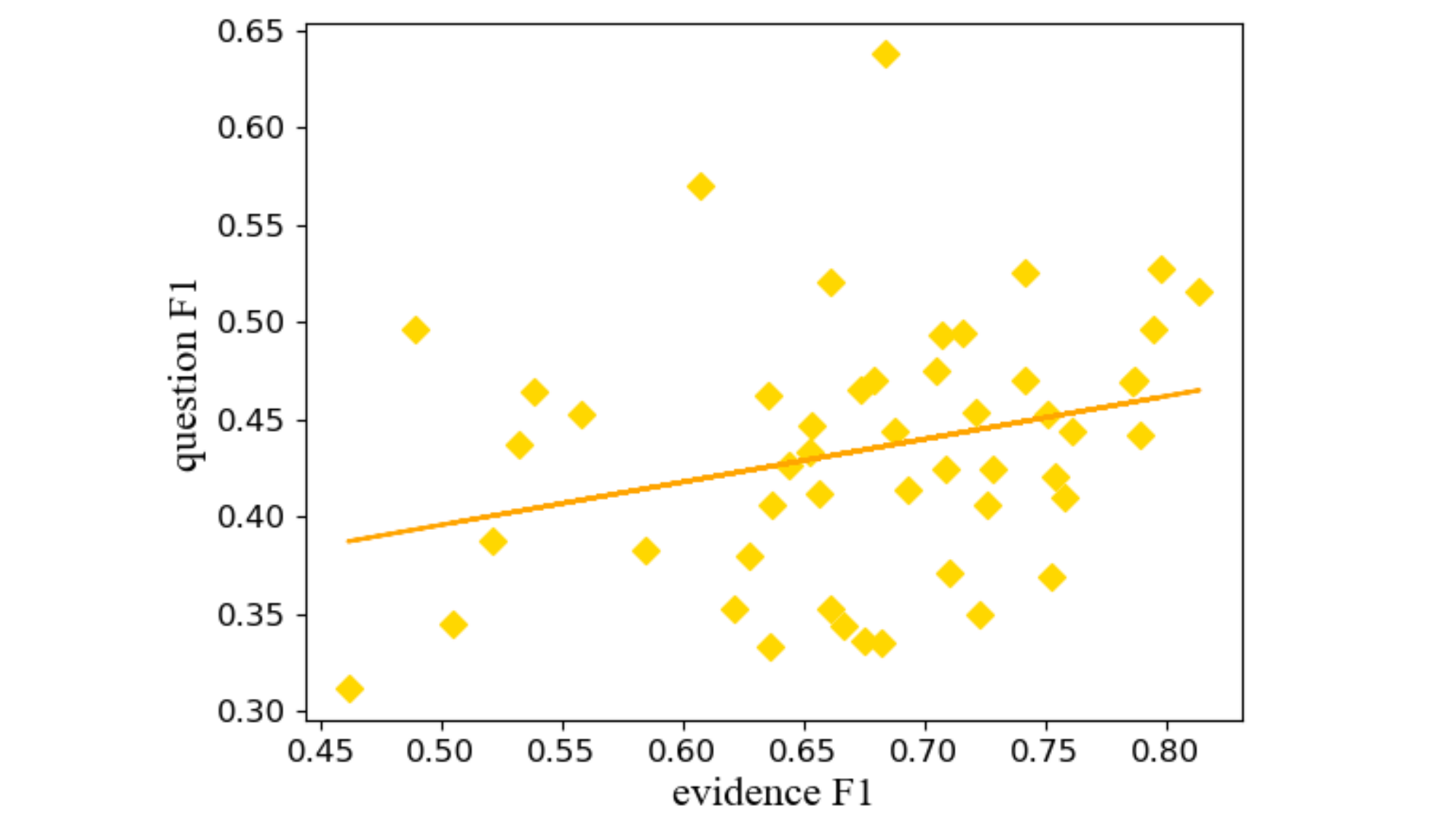}}
	 
		\centerline{QAE vs EAQ}
	\end{minipage}
 
	\caption{Performance relevance between 3 modules in our method with 13B backbone. QEA denotes evidence-aware question answering, EAQ denotes evidence-grounded query restoration and QAE denotes answer-aware evidence retrieval.}
	\label{coeff}
\end{figure*}

\begin{table}
\centering
\begin{tabular}{lcccc}
\hline Group & 1 & 2 & 3 & 4\\
\hline 
Length & 379 & 486 & 587 & 726 \\
LLama2 & 88.3 & 90.7 & 82.9 & 87.8 \\
EATQA & 90.5 &  91.9 & 86.4 & 89.3  \\ 
\hline
\end{tabular}
\caption{Results on MultiRC dataset grouped by different document lengths. Groups are indexed by the ascending order of document length, i.e., Group 1 denotes cases in the percentile interval 0-0.25 of the full dataset and Group 4 denotes cases in the percentile interval 0.75-1.0. Therefore, groups 3 and 4 have longer documents than groups 1 and 2. ``length'' denotes the average document length in the specific percentile interval. We utilize F1 to evaluate the model performance.}
\label{length}
\end{table}

\subsection{Ablation}
In this part, we investigate the effectiveness of different modules in our method, including 
QA->E, EA->Q and the distribution bridging.

\paragraph{Does question restoration matter?}
In this ablation, we remove the module of question restoration and investigate its effect on question answering. In table \ref{ablation}, removing question restoration will drop 1.7 EM and 1.0 F1 with 7B model, as well as 1.3 EM and 1.1 F1 with 13B model. Considering the context is not inputted into model in the query restoration module, the model has to utilize the information in evidence to recover the question. This module enhances the ability to arrange multiple pieces of information in evidence sentences, and understand logical relation between query, answer and evidence for LLM, which shows the effectiveness for GQA.

\paragraph{Does evidence generation matter?}
In this ablation, we remove the module of evidence generation and investigate its effect on GQA. 
In Table \ref{ablation}, removing evidence restoration will drop 2.1 EM and 0.4 F1 with 7B model, as well as 1.0 EM and 1.2 F1 with 13B model. Evidence extraction encourages the model to reason for the supporting facts that entail the question-answer pair, which enhances the understanding of logical relation among query, answer and evidence. Removing evidence generation decreases the attention of model pays to the important facts in the document.  

\paragraph{Should we narrow down the distance between $P(a,q)$ and $q(a|e,q)$?}
In this ablation, we remove the KL-divergence loss in Eq.\ref{kl} in training. In inference stage, we input the predicted evidence and the query to derive the answer.
In Table \ref{ablation}, removing KL loss will drop 0.9 EM and 0.5 F1 with 7B model, as well as 0.9 EM and 0.7 F1 with 13B model. Though keeping effective performance, the distribution bridging distills the knowledge of evidence and narrows down the gap between training and inference, avoiding first retrieving the evidence and then inputting the evidence alongside the query into model to reason for the answer. 

\begin{table}
\centering
\begin{tabular}{lcccc}
\hline Group & 1 & 2 & 3 & 4\\
\hline 
number & 10.8 & 13.5 & 16.0 & 18.3 \\
LLama2 & 87.5 & 85.1 &  85.8 & 89.0 \\
EATQA & 90.7 & 84.3 & 89.2 & 91.7  \\ 
\hline
\end{tabular}
\caption{Results on MultiRC dataset grouped by different sentence numbers in the document. Groups are indexed by the ascending order of sentence number. `number'' denotes the average sentence number in the specific percentile interval. We utilize F1 to evaluate the model performance.}
\label{number}
\end{table}

\subsection{Different document lengths and sentence number}
In this part, we investigate our performance on cases with different document lengths and sentence numbers comparing with the backbone.
We classify the MultiRC development set into 4 groups based on the document length and sentence number respectively and apply F1 to evaluate the performance of different models.

\begin{figure}[h]
	
	\begin{minipage}{0.48\linewidth}
\centerline{\includegraphics[width=\textwidth]{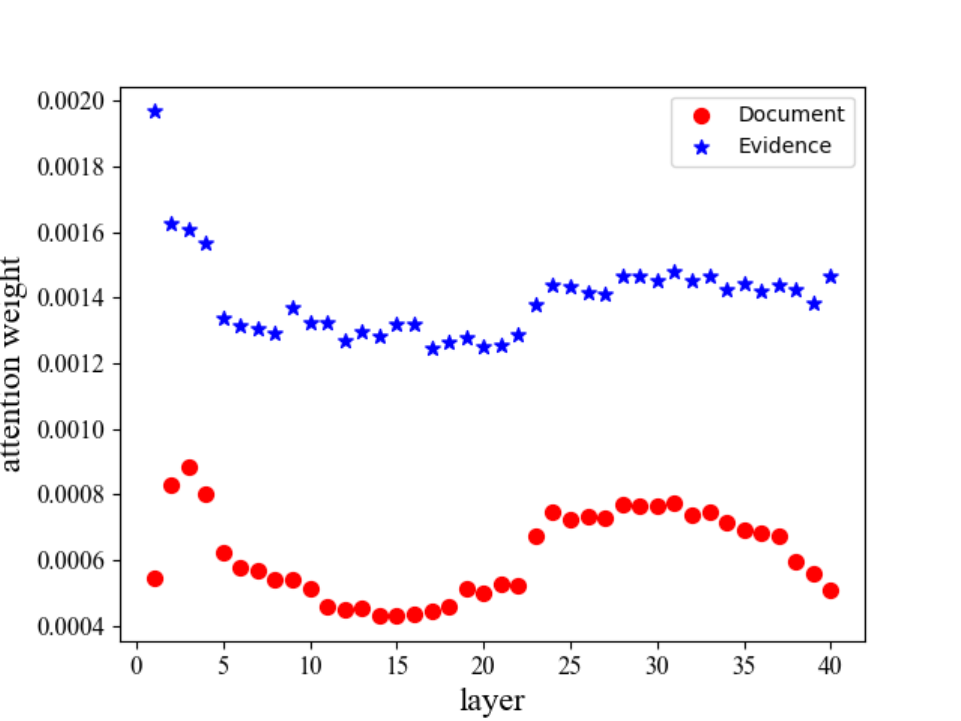}}
		\centerline{}
	\end{minipage}
	\begin{minipage}{0.48\linewidth}
\centerline{\includegraphics[width=\textwidth]{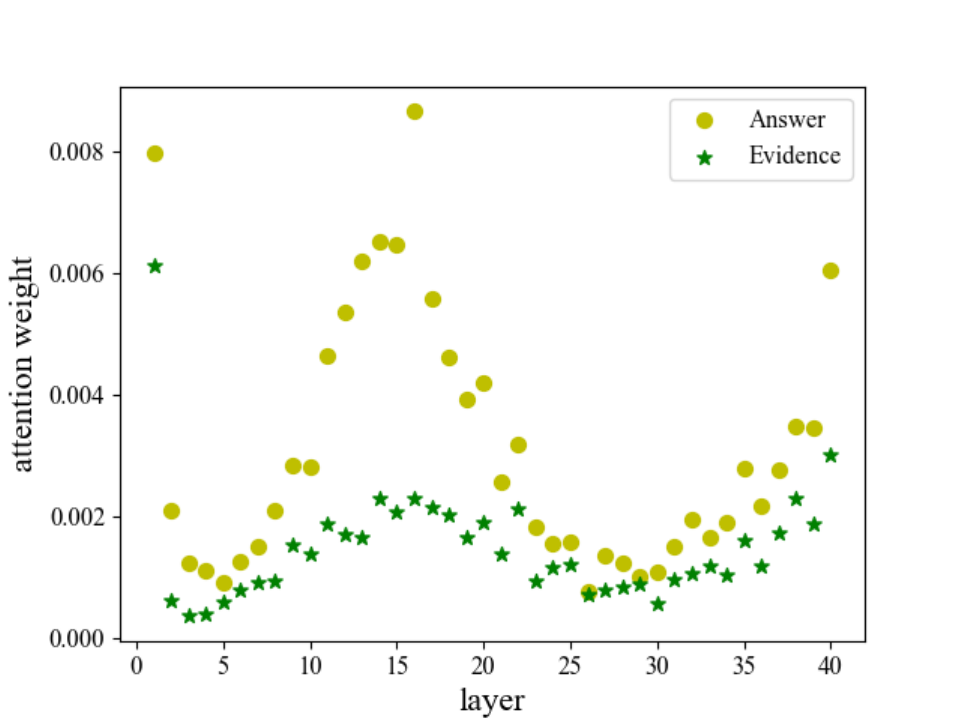}}
		\centerline{}
	\end{minipage}
	\caption{Attention weights about different layers with 13B backbone. The left graph denotes the attention weights of query to document and evidence in Evidence-Enhanced Question Answering stage; the right denotes the attention weights of generated query to evidence and answer in Evidence-Aware Question Restoration stage. }
	\label{atte}
\end{figure}


Generally, our model derives significant improvement over LLama2-13B in groups with different document lengths and sentence numbers. It demonstrates the effectiveness of our evidence enhanced triplet generation framework on document-based GQA. In Table \ref{length}, EATQA outperforms LLama2 by 3.5 and 1.5 F1 in groups 3 and 4, as well as 1.8 and 1.2 F1 in groups 1 and 2. In Table \ref{number}, EATQA outperforms LLama2 by 3.4 and 2.7 F1 in groups 3 and 4. Longer context brings the difficulty for model to capture important information about the query and derive the correct answer. Our method enhances the logical relation between evidence, query and answer, which mitigates the hallucination about distracting information in the document.


\subsection{Performance on Evidence Generation}
\begin{table}
\centering
\begin{tabular}{lcc}
\hline 
Model & 7B & 13B \\
\hline
LLama2 & 59.8 & 62.7 \\
Joint decoding & 60.3 & 63.1  \\
EATQA & \textbf{63.4} &  \textbf{65.6}  \\ 
\hline
\end{tabular}
\caption{Performance on evidence generation in MultiRC dataset. We utilize token-level F1 score as the evaluation metric. ``LLama2'' denotes instructing the LLM to generate the evidence only. ``Joint Decoding'' denotes sequentially generating evidence and answer. }
\label{evidence}
\end{table}

Not only deriving effectiveness on GQA, our method also shows improvement on evidence generation. In Table \ref{evidence}, comparing with sequentially generating evidence and answer, our method outperforms by 3.1 on 7B and 2.5 F1 on 13B. It shows our model captures the logical relation of query, evidence and answer in the unified generation framework.

\subsection{Hallucination Mitigation}
Considering the prior knowledge within LLM, we observe for some ``already-known'' questions, the model can generate the correct answer without the document, such as ``What is gravity's role in space?''. We utilize $P(Y_{A|Q} = \hat{Y})$ to evaluate the internal knowledge of model. When the model can not generate the correct answer without the document, the model resorts to the document rather than internal knowledge. We utilize the probability of generating faithful answer based on the document (without hallucinations) $P(Y_{A|Q,D} = \hat{Y} | Y_{A|Q} \neq  \hat{Y})$ to evaluate the hallucination mitigation of the model.  
In Table \ref{hallucination}, our model significantly mitigates the hallucination while keeping prior knowledge to solve the ``already-known'' questions.

\subsection{Correlation between Different Modules}

In this part, we investigate the correlation of model performance in query answering (QEA), evidence generation (QAE) and query restoration (EAQ) on data samples. To mitigate the bias of extreme sample, we classify the samples in development set into 50 groups with same size based on the QEA F1. We take the average F1 score of all samples in the group as its overall F1 score. We respectively draw the scatter plot of each pair of QEA, QAE, EAQ score versus the other and fit with linear function. In Figure \ref{coeff}. we find the QAE score and EAQ score are directly proportional to QEA score. In our triplet generation framework, with better performance in evidence generation and query restoration, the model derives better performance in query answering. This shows the effectiveness of our EATQA, which enhances the understanding of LLM
about logical relations between query, evidence and answer.

\subsection{Attention Weights}
In this part, we compute the average attention weights about query to document and evidence in the query answering task in respective layers of the transformer block. We conduct statistics on the development set of the MultiRC dataset with 13B model. In evidence-aware query answering, the model assigns about 2 times attention weights to evidence token than context token. It shows the evidence contains denser information to derive the answer. Our introduction of distribution bridging distills the abundant information in evidence to evidence-absent query answering in inference phrase. In the EA->Q, the token-average attention weights of generated query paid to evidence are comparable to answer texts. Considering the evidence contains more tokens than the answer, It shows the evidence sentences are crucial for the feasibility of EA->Q task.

\section{Conclusion}
In this paper, we propose the unified triplet generation framework including three instruction tuning tasks to improve the logical reasoning ability of LLM for GQA task. 
We conduct experiments on two widespread document-based QA datasets MultiRC and QASPER with different sizes of LLM, and achieve the new state-of-the-art on both datasets.

\section*{Limitations}
In this paper, we propose the unified triplet generation framework to improve the logical reasoning ability of LLM for GQA task. We conduct experiments with different sizes of LLama and achieve the new SOTA. We will conduct experiments on other LLMs with different architectures in further research. 
\bibliography{anthology,custom}
\bibliographystyle{acl_natbib}

\appendix

\section{Appendix}
\label{sec:appendix}

Input templates of different modules in EATQA. 
\begin{figure}[h]
    \centering
\includegraphics[width=0.5\textwidth]{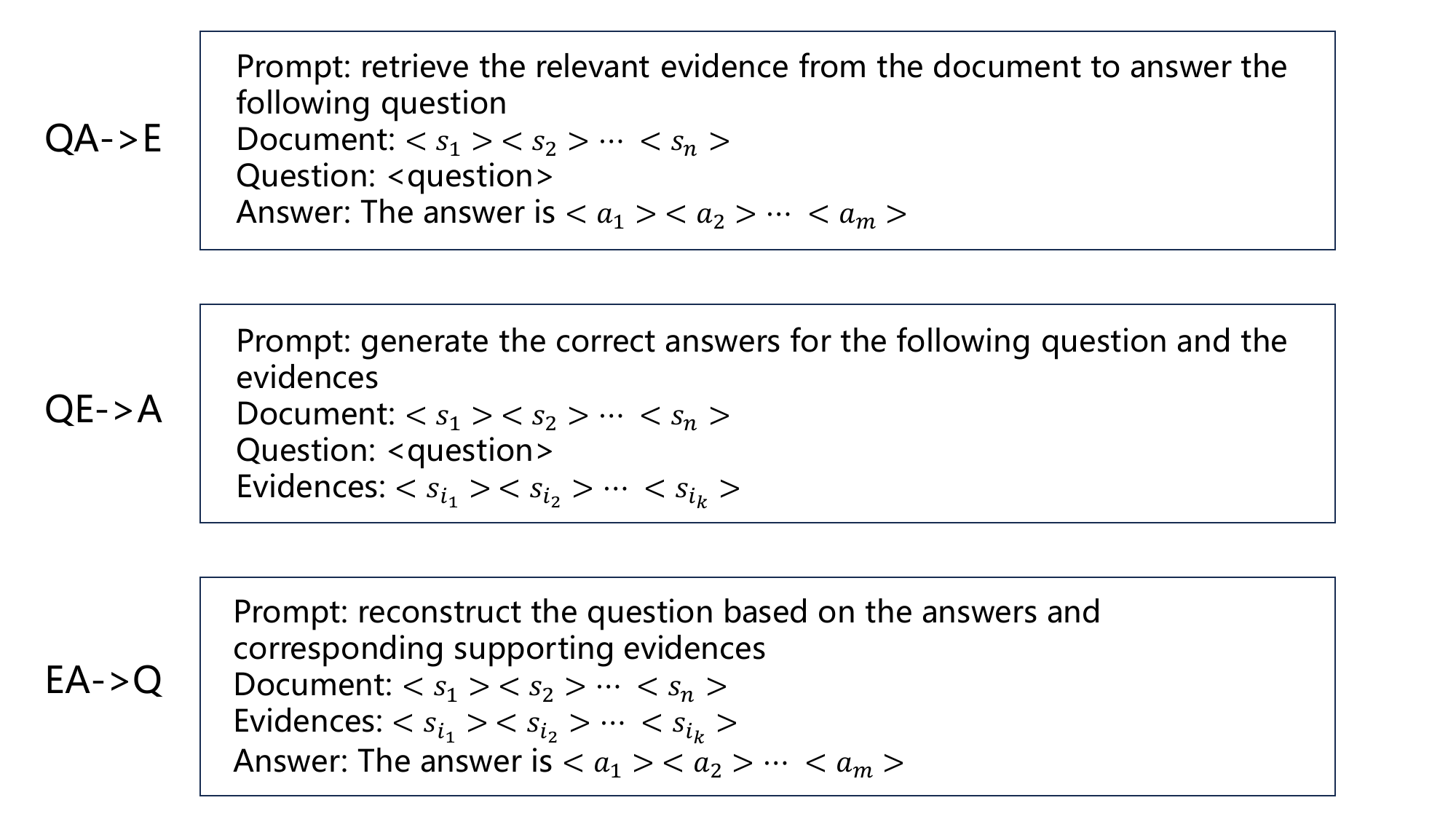}
    \caption{Input templates of EATQA.}
    \label{temp}
\end{figure}
\end{document}